# High Performance Human Face Recognition using Gabor based Pseudo Hidden Markov Model

Arindam Kar[1], Debotosh Bhattacharjee[2], Mita Nasipuri[2], Dipak Kumar Basu[2*], Mahantapas Kundu[2]
[1] Indian Statistical Institute, Kolkata-700108, India
[2] Department of Computer Science and Engineering, Jadavpur University, Kolkata- 700032, India
* Former Professor & AICTE Emeritus Fellow
Email: {kgparindamkar@gmail.com, debotosh@indiatimes.com, m.nasipuri@ cse.jdvu.ac.in, dipakkbasu@gmail.com, mkundu@cse.jdvu.ac.in}

*Abstract*— This paper introduces a novel methodology that combines the multi-resolution feature of the Gabor wavelet transformation (GWT) with the local interactions of the facial structures expressed through the Pseudo Hidden Markov model (PHMM). Unlike the traditional zigzag scanning method for feature extraction a continuous scanning method from top-left corner to right then top-down and right to left and so on until right-bottom of the image i.e. a spiral scanning technique has been proposed for better feature selection. Unlike traditional HMMs, the proposed PHMM does not perform the state conditional independence of the visible observation sequence assumption. This is achieved via the concept of local structures introduced by the PHMM used to extract facial bands and automatically select the most informative features of a face image. Thus, the long-range dependency problem inherent to traditional HMMs has been drastically reduced. Again with the use of most informative pixels rather than the whole image makes the proposed method reasonably faster for face recognition. This method has been successfully tested on frontal face images from the ORL, FRAV2D and FERET face databases where the images vary in pose, illumination, expression, and scale. The FERET data set contains 2200 frontal face images of 200 subjects, while the FRAV2D data set consists of 1100 images of 100 subjects and the full ORL database is considered. The results reported in this application are far better than the recent and most referred systems.

## I. Introduction

Face recognition is one of the major topics in the image processing and pattern recognition research area. The applications of face recognition systems are manifold, like access control, video surveillance, credit card user identification and automatic video indexing. In recent years many approaches to face recognition have been developed. [1, 2] give an overview of the different face recognition techniques.

Hidden Markov Model (HMM) is a very important methodology for modelling structures and sequence analysis. It mostly involves local interaction modeling. With the current perceived world security situation, governments, as well as businesses, require reliable methods to accurately identify individuals, without overly infringing on rights to privacy or requiring significant compliance on the part of the individual being recognized. Person recognition systems based on biometrics have been used for a significant period for law enforcement and secure access. Both fingerprint and iris recognition systems are proven as reliable techniques; however, the method of capture for both limits their versatility [3]. Face recognition is a complicated task; this is due to the increased variability of acquired face images [4]. Controls can sometimes be placed on face image acquisition, for example, in the case of passport photographs; but in many cases this is not possible. Variations in pose, expression, illumination and partial occlusion of the face therefore become nontrivial issues that have to be addressed.

Several systems use Hidden Markov Models for face recognition [1, 5, 6, 7]. In this paper, focus has been given on Pseudo Hidden Markov Model based face recognition system. The proposed Gabor based Pseudo Hidden Markov Model (PHMM) approach allows both the structural and the statistical properties of a pattern to be represented within the same probabilistic framework. This approach also allows the user to weight substantially the local structures within a pattern that are difficult to disguise. This provides a PHMM recognizer with a higher degree of robustness. Indeed, PHMMs have been shown to outperform HMMs in a number of applications. However, PHMMs are well suited to model the inner and outer structures of any sequential pattern (such as a face) simultaneously.

As well as being used in conjunction with HMMs for face recognition, Gabor Wavelet Transformation (GWT) has been coupled with other techniques. Its ability to localize information in terms of both frequency and space (when applied to images) make it an invaluable tool for image processing. Here GWT is used to extract i) high frequency features, i.e. feature vectors are extracted at points on the face image with high information content. ii) the significant structures of the face, enabling statistical measures to be calculated as a result reinforced by selecting only those pixels which are high energized pixels of the GWT image, reduces complexity and works better with occlusions. The Gabor wavelet in particular has been used extensively for face recognition applications.

Hidden Markov models and related techniques have been applied to gesture recognition tasks with success. The technique is motivated by the works of Samaria and

Young [7, 8], and Kohir & Desai [9]. HMM models provide a high level of flexibility for modeling the structure of an observation sequence, they also allow for recovering the (hidden) structure of a sequence of observations by pairing each observation with a (hidden) state. State duration is left free so that HMM represents a powerful technique. The two-dimensional structure of the image is accounted by using a pseudo model. In this paper a structure similar to the PHMM is created with the effort the use of only the most informative features of the GWT image is investigated, and the influence of new scanning technique together with the effect of window size is also analyzed here, on the dataset taken from the ORL [10], FERET [11], and the FRAV2D [12] face databases.

## II. 2-D GABOR WAVELET TRANSFORMATION

GWT has been recognized as a powerful tool in a wide range of applications, including image/video processing, numerical analysis, and telecommunication. A Gabor wavelet is convolved with an image either locally at selected points in the image, or globally. The Gabor wavelets (kernels, filters) can be defined as follows:

$$\varphi_{\mu,\nu}(z) = \frac{|k_{\mu,\nu}|^2}{\sigma^2} e^{-\|k_{\mu,\nu}\|^2 \|z\|^2 / 2\sigma^2} \left[ e^{ik_{\mu,\nu}z} - e^{-(\sigma^2/2)} \right] \quad (1)$$

where $\mu$ and $\nu$ defines the orientation and scale of Gabor kernels respectively, $z = (x, y)$, is the variable in spatial domain, $\| \|$ denotes the norm, and $k_{\mu,\nu}$ is the frequency vector which determines the scale and orientation of Gabor kernels, $k_{\mu,\nu} = k_\nu e^{i\phi_\mu}$ where $k_\nu = k_{max}/f^\nu$, $k_{max} = \pi/2$, $\phi_\mu = \pi\mu/8$, and $f$ is the spacing factor between kernels in the frequency domain. Here we use Gabor wavelets at five different scales, $\nu \in \{0,...,4\}$, and eight orientations, $\mu \in \{0,...,7\}$. The kernels exhibit strong characteristics of spatial locality and orientation selectivity, making them a suitable choice for image feature extraction when one's goal is to derive local and discriminating features for (face) classification. It is to be noted that we applied the magnitude but did not use the phase, which is considered to be consistent with the application of Gabor wavelet representations in [13, 14]. As the outputs $(O_{\mu,\nu}(z) : \mu \in \{0,...,4\}, \nu \in \{0,...7\})$ consist of 40 different local, scale and orientation features, the dimensionality of the Gabor transformed image space is very high. So dimension reduction becomes essential for Gabor transformed image. The term $e^{-(\sigma^2/2)}$ in equation (1) is subtracted in order to make the kernel DC-free, thus becoming insensitive to illumination.

## III. GABOR BASED PSEUDO HIDDEN MARKOV MODEL FOR FACE RECOGNITION

The first approach to face recognition based on PHMM is described in [6]. In [15], it is also assessed that not only such HMM structure would lead to an unmanageable complexity but the retrieval of the best 2D state sequences is not ensured.

### A. Preprocessing for PHMM

In this approach some preprocessing is necessary on the original image before feature extraction. The preprocessing steps are as follows:
1. Convolution output of images with Gabor kernels is obtained.
2. As the convolution outputs contain both real and imaginary parts, so each pixel value of the convolution output is replaced by the $L_1$ norm of that pixel.
3. The pixels at identical positions of all the Gabor transformed image are summed up to obtain the final single Gabor transformed image and is termed as $G_F$.
Fig 1(a), Fig1(b) and Fig. 1(c) shows the original image, the 40 convolution outputs of the image and the resultant final Gabor wavelet transformed image respectively.

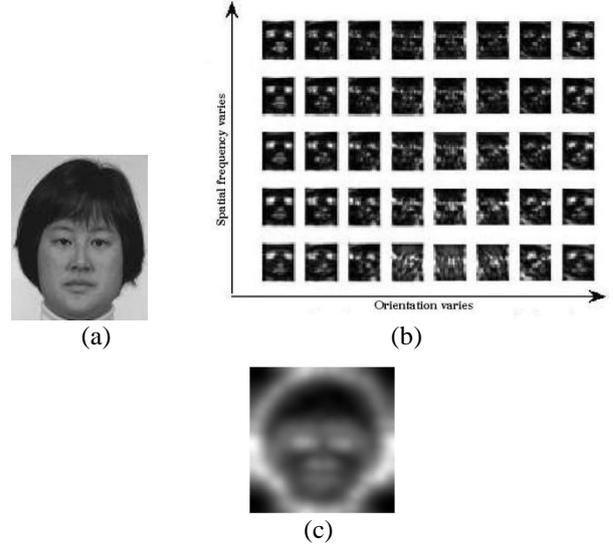

**Fig 1.** Gabor Wavelet transform of image: (a) original image, (b) 40 Convolution outputs, (c) final single Gabor transformed image.

### B. Preparing Face Images for PHMM Analysis

HMM provides a way of modeling the one dimensional [1D] signals using their statistical properties. The 1D nature of signals, for example speech [16, 17], is well suited to analysis by HMM. Though images are two dimensional (2D) and as there exists no direct equivalent of 1D HMM for 2D signals. So attempts have been made to use multi-model representations that give pseudo 2D structure [18, 19]. This poses the problem of extracting a meaningful ID sequence from a 2D image. The solution is to consider either a temporal or a spatial sequence. In [19] it was concluded that the spatial sequence of windowed data give more meaningful models.

In this approach the final Gabor transformed image $G_F$ is divided into overlapping strips of height h pixels and length which is same as width of the image and overlap of $p$ pixels as shown in Fig.2 for feature extraction. Firstly each strip of the image is scanned using horizontal and vertical overlapping blocks using the spiral scanning technique, shown in Fig 4. An overlap between adjacent sampling blocks improves the ability of the HMM to model the neighborhood relations between the sampling blocks. Each strip is subsequently divided into blocks of size ($k \times k$) pixels, with horizontal and vertical overlap of size $p$ pixels between the blocks. This is illustrated in Fig 2 (a), (b) where $N_S$, total number of blocks of strip s, depends on the strip size, block size and the overlapping size. Thus, in an image strip of width $w$ and height $h$, there will be $\left(\frac{h-p}{k-p}\right) * \left(\frac{w-p}{k-p}\right)$ blocks. From the experimental results it is seen that the best results are achieved with block size of 16×16 and an overlap of 75 % in each direction. Fig. 3 shows the sampling method and the resulting array of vectors.

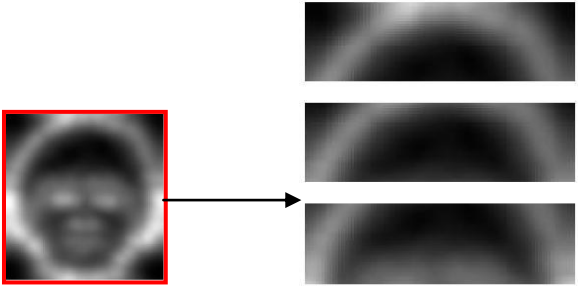

Fig. 2.(a) GWT face image being segmented into overlapping strips

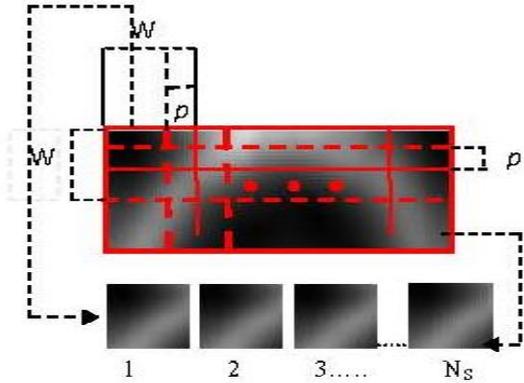

Fig. 2.(b)

**Fig 2**. (a) & (b) shows an illustration showing the creation of the block sequence.

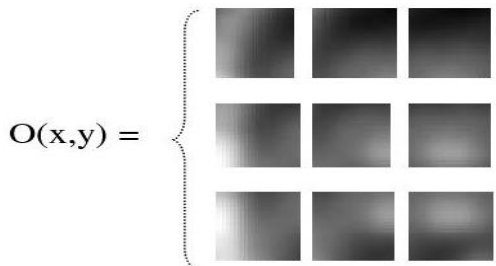

**Fig 3**. Sampling Technique

C. *Extraction of 2-D GWT Coefficients for PHMM Analysis.*

As HMMs are 1D data modelers. So for 2D image applications, they have to be redefined in 2D context, or 2D data is converted to 1D, Here, with a new approach 2D face image data is converted to 1D sequence, without losing significant information and this 1D sequence is used for HMM training and testing. The features are modeled by PHMM. These models are extensions of the HMM. In this paper, the observation vector is constructed, with the most informative feature being collected from each block. To encompass the different spatial frequencies (scales), spatial localities, and orientation selectivity, all the Gabor wavelet responses are concatenated to obtain a single matrix $G_F$.

Derivation of the observation vector $O_i$ is as follows:

Start

1. Find $\bar{g}_\alpha$ = [mean of ($G_F$)].
2. For each block B of size $k \times k$ pixels, find pixels (x, y) in B such that $G_F(x, y) \geq \bar{g}_\alpha$, then consider the sum of all the pixels that satisfy the above criteria consider the sum as $S = \sum G_F(x, y)$.
3. If no such pixel exists then take $S = k \times \bar{g}_\alpha$, where k is a constant, here k is taken as the block size.
4. Store S i.e., the feature points into the column vector $O_i$.
5. Repeat step 2 to step 4 for all the blocks.

End

At this stage the 2-D face image is represented by a 1D GWT domain observation sequence, which is truncated to a lower dimension than the original image. Only the most informative pixels of the GWT image are considered for creation of the observation vector, make it suitable for HMM training and testing, and hence make the computation faster.

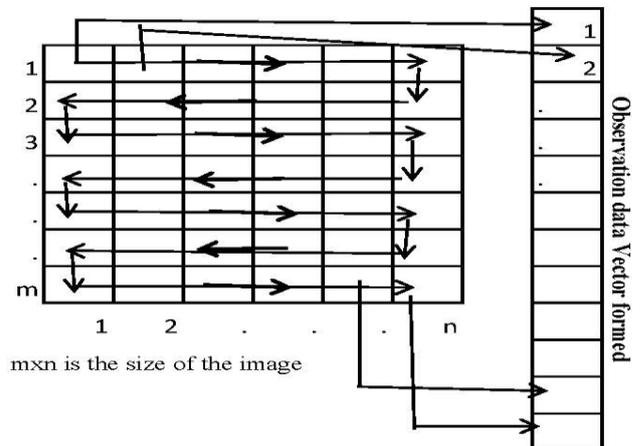

**Fig 4.** Technique for converting GWT coefficients to 1D vector.

IV. IMPLEMENTATION OF PHMM FOR RECOGNITION

The obtained 1D observation vector from all the training images of an individual is used for HMM training using the Baum-Welch algorithm. The size of these observation vectors depends on the size of the sliding block, the overlap size between the adjacent blocks and the size of the image strip taken. The block size has to be chosen in such a way that it covers most significant coefficients, i.e., the coefficients that contain most information are extracted. The sampling technique converts each image into a sequence of single column vector, which is spatially ordered in a top-to-bottom sense is described in Fig. 2. Each vector O represents the most informative intensity level values of pixels contained in the corresponding block. Assuming that each face is in an upright, frontal position, feature regions occurs in a predictable order, for example eyes following forehead, nose, lips, and so on. This ordering suggests the use of a cyclic HMM (see figure 5 for a 7-state model), where only transitions between adjacent states in a single direction (cyclic manner) is allowed. A PHMM model can be defined by the following elements.

(i) N is the number of hidden states in the model.
(ii) M is the number of observation features.
(iii) $S = \{S_1, S_2, \ldots, S_N\}$ is the finite set of possible hidden states. The state of the model at time $t$ is given by $q_t \in S$, $1 \leq t \leq T$, where $T$ is the length of the observation sequence.
(iv) $A = \{a_{ij}\}$ is the state transition probability matrix where $a_{ij} = P[q_{t+1} = S_j / q_t = S_i], 1 \leq i, j \leq N$ with $0 \leq a_{ij} \leq 1$ and $\sum_{j=1}^{N} a_{ij} = 1$, $1 \leq i \leq N$. This cyclic HMM will segment the image into statistically similar regions, each of which is represented by a facial band.

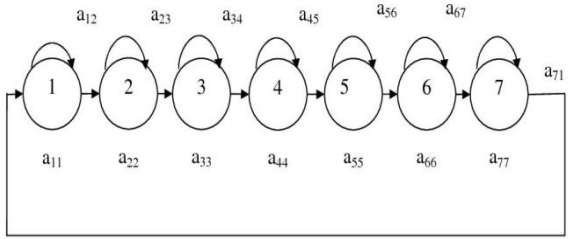

**Fig 5.** Cyclic Hidden Markov Model

V. IMPROVEMENT OF THIS SYSTEM

Although this system is based on PHMM, too, but the proposed method achieves much higher recognition rates than the other previous systems. In this section we will describe the differences of our system to the other previous described system and evaluate the effects on the recognition rate.

A. *Minimizing the effect of illumination*

To the best of the literature we have surveyed in other works the feature extraction in done by scanning the original image. But in this approach feature extraction is done by scanning the GWT image instead of the original image, which is less sensitive to the overall level of illumination. Hence the overall effect of the illumination is minimized.

B. *Most informative Feature extraction*

Another major improvement of our system is the use of only few most informative Gabor transformed coefficients as features extracted by the method shown in section 3.2B rather than the whole image. As the Gabor wavelet representation captures salient visual properties such as spatial localization, orientation selectivity, and spatial frequency, and it is known that the high frequency components of Gabor wavelet transformation provides finer details needed in the identification process [20]. So these extracted features by the proposed feature extraction method further enhance the face recognition performance in presence of occlusions. Further, as only the high energized features are extracted from each block it reduces the length of the observation vector and hence reduces computational complexity. We consider this as one of the main improvements of our recognition system.

C. *Spiral Scanning Technique*

Experiment conducted shows that on the extracted most informative pixels as features, by the spiral scanning technique as shown in Fig. 4, achieved nearly 5% higher recognition rate than the standard zigzag scanning. We think that this has advantages for the recognition as for eg. When there is a change in pose or facial expression important parts in the human face, for e.g. eyes are not in the same state on the same level in the image for the zigzag scanning. This effect has been compensated by scanning the blocks in continuous order by the states. Again the regions of the face are aligned to the same state. The extraction is more accurate than the extraction of features in [7].

VI. SIMILARITY MEASURES AND CLASSIFICATION

The optimal paths obtained from the PHMM are the representative of the images. The minimum distances between the optimal paths of the images are used for image classification. Let $M'_k$ be the mean of the training samples for class $w_k$, where $k = 1,2,\ldots,l$ where $l$ is the number of classes. The classifier then applies, the nearest neighbor (to the mean) rule for classification using the similarity (distance) measure $\delta$ :

$$\delta(Y, M'_j) = \min_k (\delta(Y, M'_k)) \Rightarrow Y \in w_j \quad (2)$$

The optimal path vector $Y_{opt}$ is classified to that class of the closest mean $M'_k$ using the similarity measure $\delta$. The similarity measures used here are, $L_2$ distance measure, $\delta_{L_2}$, Mahalanobis distance measure $\delta_{Md}$ [21], and the cosine similarity measure, $\delta_{\cos}$ [22], which are defined as:

$$\delta_{Md} = (X - Y)^t \Sigma^{-1} (X - Y), \quad (3)$$

$$\delta_{\cos} = \frac{-X^t Y}{\|X\| \|Y\|}, \quad (4)$$

$$\delta_{L_2} = (X - Y)^t (X - Y) \quad (5)$$

here $\Sigma$ is the covariance matrix; $\|\ \|$ is the $L_2$ norm operator; and t is the transpose operator.

## VI. EXPERIMENT ON THE FRONTAL AND POSE-ANGLED IMAGES FOR FACE RECOGNITION

This section assesses the performance of the Gabor-based PHMM method for both frontal and pose-angled face recognitions. The effectiveness of the Gabor-based PHMM method has been successfully tested on the dataset taken from the FRAV2D FERET, and ORL face databases. For frontal face recognition, the data set is taken from the FRAV2D and ORL database. For pose-angled face recognition, the data set is taken from the FERET database.

### A. FERET Database

The FERET database, employed in the experiment here contains, 2,200 facial images corresponding to 200 individuals with each individual contributing 11 images. The images in this database were captured under various illuminations and display, a variety of facial expressions and poses. As the images include the background and the body chest region, so each image is cropped to exclude those, and then scaled to $92 \times 112$. Fig. 6 shows all samples of one subject. The details of the images are as follows: (a) regular facial status; (b) +15° pose angle; (c) -15° pose angle; (d) +25° pose angle; (e) -25° pose angle; (f) +40° angle; (g) -40° pose angle; (h) +60° pose angle; (i) -60° pose angle; (j) alternative expression; (k) different illumination.

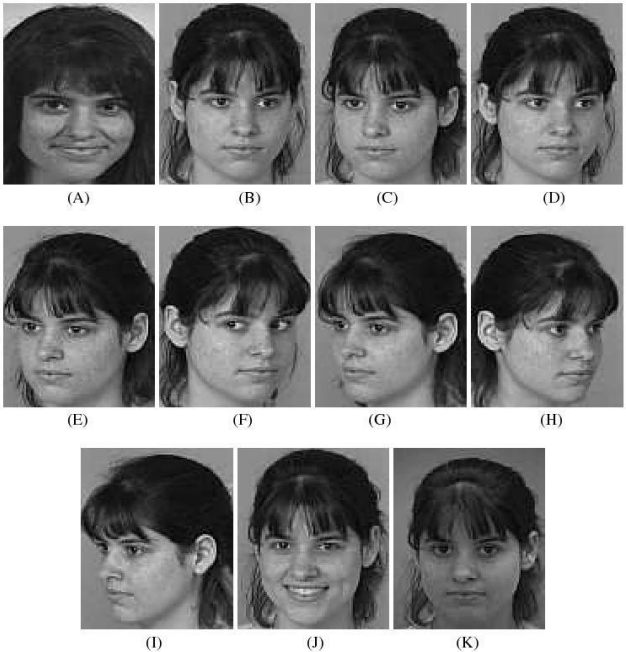

**Fig 6.** Demonstration images of an individual from the FERET Database

Table 1. Average recognition results using FERET database:

| Method | Recognition Rates (%) No. of training samples | | Avg. Recognition Rates (%) |
|---|---|---|---|
| | 3 | 4 | |
| GWT | 79 | 83.5 | 81.25 |
| PHMM | 82.25 | 86.5 | 84.375 |
| **GWT PHMM** | **88.5** | **91.5** | **90** |

### B. Sensitivity and Specificity measures of the FERET dataset

To measure the sensitivity and specificity [23], the dataset from the FERET database is prepared in the following manner. A total 200 class is made, from the dataset of 2200 images of 200 individuals where each class consists of 18 images in it. Out of the 18 images in each class, 11 images are of a particular individual, and 7 images are of other individual taken using permutation as shown in Fig. 7. From the 11 images of the particular individual, at first the first 4 images (A-D), then the first 3 images (A-C) of a particular individual are selected as training samples and the remaining images of the particular individual are used as positive testing samples. The negative testing is done using the images of the other individual. Fig. 7 shows all sample images of one class of the data set used from FERET database.

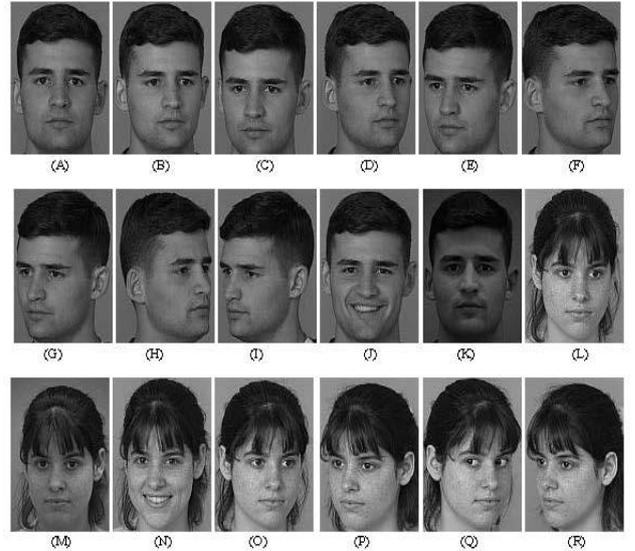

**Fig. 7.** Demonstration images of one class from the FERET database

Table 2: Specificity & Sensitivity measures of the FERET database:

| | | Total no. of classes=200, Total no. of images= 3600 | |
|---|---|---|---|
| | | Individual belonging to a particular class | |
| | | Using first 3 images of an individual as training images | |
| | | Positive | Negative |
| **FERET test** | Positive | $(T_P) = 1416$ | $(F_P) = 31$ |
| | Negative | $(F_N) = 184$ | $(T_N) = 1369$ |
| | | Sensitivity = $T_P$ / $(T_P + F_N)$ = 88.5% | Specificity = $T_N$ / $(F_P + T_N) \approx$ 97.78% |
| | | Using first 4 images of an individual as training images | |
| | | Positive | Negative |
| **FERET test** | Positive | $(T_P) = 1281$ | $(F_P) = 20$ |
| | Negative | $(F_N) = 119$ | $(T_N) = 1380$ |
| | | Sensitivity = $T_P$ / $(T_P + F_N)$ = 91.5% | Specificity = $T_N$ / $(F_P + T_N) \approx$ 98.57% |

Thus considering **first 3 images (A-C)** of a particular individual for training the achieved rates are:
**False positive rate = $F_P / (F_P + T_N)$ = 1 − Specificity =2.22%
False negative rate = $F_N / (T_P + F_N)$ = 1 − Sensitivity=11.5%
Accuracy = $(T_P+T_N)/(T_P+T_N+F_P+F_N)$ ≈93.1.**
Considering first **4 images (A-D)** of a particular individual for training the achieved rates are:
**False positive rate = $F_P / (F_P + T_N)$ = 1 − Specificity =1.43%
False negative rate = $F_N / (T_P + F_N)$ = 1 − Sensitivity=8.50%
Accuracy = $(T_P+T_N)/(T_P+T_N+F_P+F_N)$ ≈95.1.**

*C. FRAV2D Database*

The FRAV2D face database, employed in the experiment consists; 1100 colour face images of 100 individuals, 11 images of each individual are taken, including frontal views of faces with different facial expressions, under different lighting conditions. All colour images are transformed into gray images and scaled to 92×112. Fig. 8 shows all samples of one individual. The details of the images are as follows: (A) regular facial status; (B) and (C) are images with a 15° turn with respect to the camera axis; (D) and (E) are images with a 30° turn with respect to the camera axis; (F) and (G) are images with gestures; (H) and (I) are images with occluded face features; (J) and (K) are images with change of illumination.

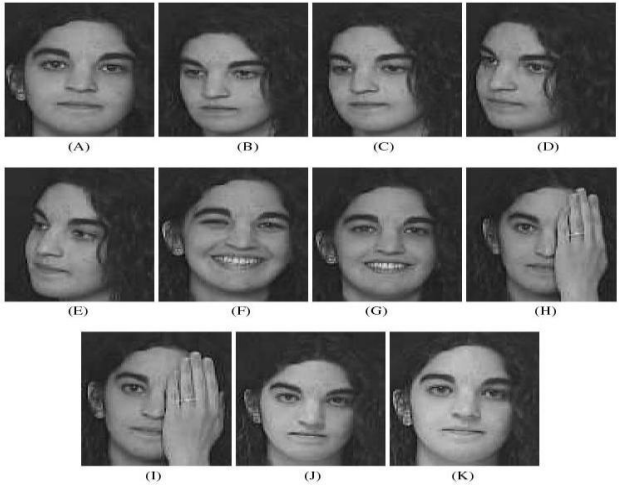

**Fig 8**. Demonstration images of one subject from the FRAV2D database

Table 3. Average recognition results using FRAV2D database:

| Method | Recognition Rates(%) No. of training samples | | Avg. Recognition Rates (%) |
|---|---|---|---|
| | 3 | 4 | |
| GWT | 81.5 | 84.75 | 83.25 |
| PHMM | 85 | 90.75 | 87.875 |
| **GWT PHMM** | **94.5** | **96.9** | **95.7** |

*D. Sensitivity and Specificity measures of the FRAV2D dataset*

To measure the sensitivity and specificity, the dataset from the FRAV2D database is prepared in the following manner. A total 100 class is made, from the dataset of 1100 images of 100 individuals where each class consists of 18 images in it. Out of the 18 images in each class, 11 images are of a particular individual, and 7 images are of other individual taken using permutation as shown in Fig. 9. From the 11 images of the particular individual, at first the first 4 images (A-D), then the first 3 images (A-C) of a particular individual are selected as training samples and the remaining images of the particular individual are used as positive testing samples. The negative testing is done using the images of the other individual. Fig. 9 shows all sample images of one class of the data set used from FRAV2D database.

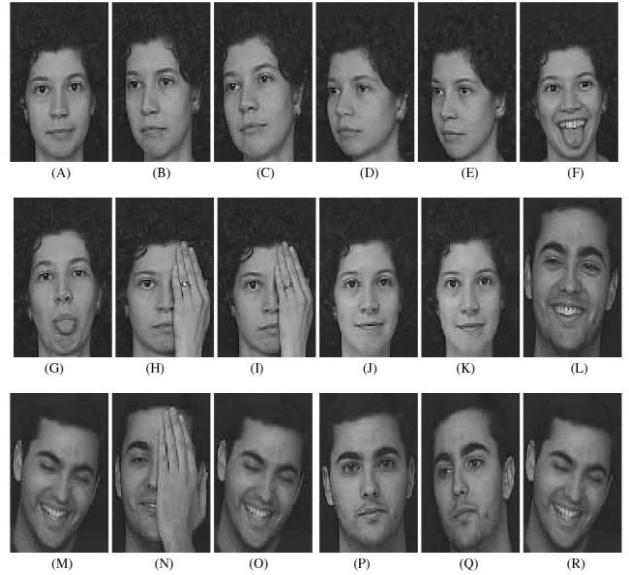

**Fig 9**. Demonstration images of one class from the FRAV2D database

Table 4: Specificity & Sensitivity measures of the FRAV2D database:

| | | Total no. of classes=100, Total no. of images= 1800 | |
|---|---|---|---|
| | | **Individual belonging to a particular class** | |
| | | **Using first 3 images of an individual as training images** | |
| | | Positive | Negative |
| **FRAV2D test** | **Positive** | $(T_P) = 756$ | $(F_P) = 7$ |
| | **Negative** | $(F_N) = 44$ | $(T_N) = 693$ |
| | | Sensitivity = $T_P / (T_P + F_N)$ ≈ 94.5% | Specificity =$T_N/(F_P + T_N)$ =99.0% |
| | | **Using first 4 images of an individual as training images** | |
| | | Positive | Negative |
| **FRAV2D test** | **Positive** | $(T_P) = 678$ | $(F_P) = 2$ |
| | **Negative** | $(F_N) = 22$ | $(T_N) = 698$ |
| | | Sensitivity = $T_P / (T_P + F_N)$ ≈ 96.9% | Specificity =$T_N/(F_P + T_N)$ ≈99.7% |

Thus considering the **first 3 images (A-C)** of a particular individual for training the achieved rates are:
False positive rate = $F_P / (F_P + T_N)$ = 1 − Specificity =1.0%
False negative rate = $F_N / (T_P + F_N)$ = 1 − Sensitivity=5.5%
**Accuracy = $(T_P+T_N)/(T_P+T_N+F_P+F_N)$ ≈96.8.**
Considering the **first 4 images (A-D)** of a particular individual for training the achieved rates are:
False positive rate = $F_P / (F_P + T_N)$ = 1 − Specificity =.3%
False negative rate = $F_N / (T_P + F_N)$ = 1 − Sensitivity=2.1%
**Accuracy = $(T_P+T_N)/(T_P+T_N+F_P+F_N)$ ≈98.3.**

*E. ORL Database*

The whole ORL database, is considered here, each image is scaled to $92 \times 112$ with 256 gray levels. Fig. 10 shows all samples of one individual from the ORL database.

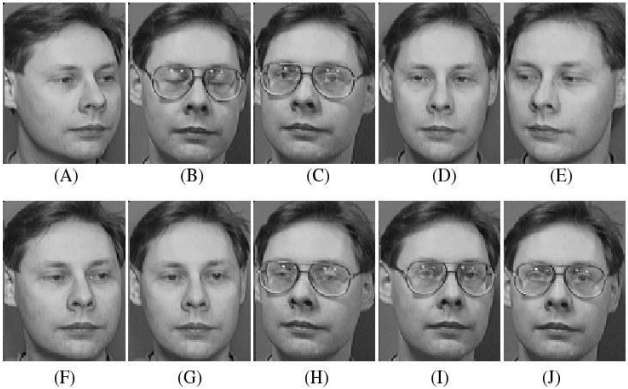

**Fig 10.** Demonstration images of one subject from the ORL Database

*F. Sensitivity and Specificity measures of the ORL database*

To measure the sensitivity and specificity, the dataset from the ORL database is prepared in the following manner. A total 40 class is made, from the dataset of 400 images of 40 individuals where each class consists of 15 images in it. Out of the 15 images in each class, 10 images are of a particular individual, and 5 images are of other individual taken using permutation as shown in Fig. 11. From the 10 images of the particular individual, at first the first 2 images (A-B), then only the first image (A) of a particular individual are selected as training samples and the remaining images of the particular individual are used as positive testing samples. The negative testing is done using the images of the other individual. Fig. 11 shows all sample images of one class of the data set taken from ORLET database.

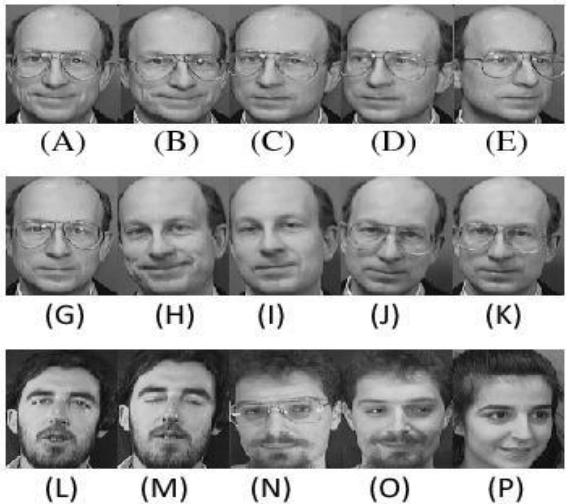

**Fig 11.** Demonstration images of one class from the ORL database.

Table 5: Specificity & Sensitivity measures of the ORL database:

| | | Total no. of classes=40, Total no. of images= 600 | |
|---|---|---|---|
| | | Individual belonging to a particular class | |
| | | Using only the first image of an individual as training image | |
| | | Positive | Negative |
| ORL test | Positive | $(T_P) = 360$ | $(F_P) = 0$ |
| | Negative | $(F_N) = 0$ | $(T_N) = 200$ |
| | | Sensitivity = $T_P / (T_P + F_N)$ = 100% | Specificity = $T_N / (F_P + T_N)$ = 100% |
| | | Using first 2 images of an individual as training images | |
| | | Positive | Negative |
| ORL test | Positive | $(T_P) = 320$ | $(F_P) = 0$ |
| | Negative | $(F_N) = 0$ | $(T_N) = 200$ |
| | | Sensitivity = $T_P / (T_P + F_N)$ = 100% | Specificity = $T_N / (F_P + T_N)$ = 100% |

Thus considering only the **first image (A)** of a particular individual for training the achieved rates are:
False positive rate = $F_P / (F_P + T_N) = 1 -$ Specificity = 0%
False negative rate = $F_N / (T_P + F_N) = 1 -$ Sensitivity = 0%
**Accuracy = $(T_P+T_N)/(T_P+T_N+F_P+F_N)$ =100.**
Considering the **first 2 images (A-B)** of a particular individual for training the achieved rates are:
False positive rate = $F_P / (F_P + T_N) = 1 -$ Specificity = 0%
False negative rate = $F_N / (T_P + F_N) = 1 -$ Sensitivity = 0%
**Accuracy = $(T_P+T_N)/(T_P+T_N+F_P+F_N)$ =100.**
The obtained result by applying the proposed approach on the ORL database images outperforms, most of the other previous methods are shown in table 6.

Table 6: Comparison of recognition performance of the proposed Gabor based PHMM method with some other method on the ORL database.

| Methods | Sensitivity Rate(%) | Reference |
|---|---|---|
| Local Gabor wavelet | 66 | Huang et al. (2004) [24] |
| Eigenface | 72.1 | Yin et. al. (2005) [25] |
| Fisherface | 76.3 | Yin et. al. (2005) [25] |
| PCA | 82.8 | Huang et al. (2004) [24] |
| ICA | 85 | Huang et al. (2004) [24] |
| Ergodic HMM +DCT | 99.5 | Kohir and Desai (1998) [9] |
| 2D PHMM + neural network coefficients | 100 | Vitoantonio Bevilacqua et al (2007) [26] |
| **Proposed method** | **100** | **This paper** |

*VII. EXPERIMENTAL RESULTS*

Fig. 12, 13, and Fig. 14 shows the positive face recognition performance i.e. the sensitivity of the Gabor based PHMM method; with the same dataset by varying the size of blocks using four different similarity measures on the dataset from the FERET, FRAV2D and ORL database respectively. The horizontal axis indicates the blocks size used, and the vertical axis represents the positive face recognition rate i.e. the sensitivity, which is the rate that the top response is correct (in the correct class).

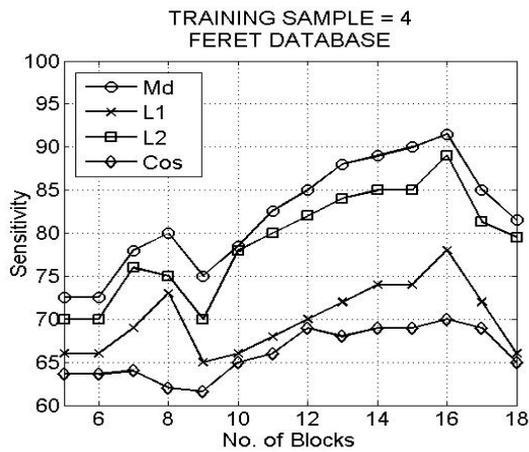

Fig.12. Sensitivity measure of the Gabor based PHMM method on the FERET database, using the four different similarity $M_d$ (the Mahalanobis distance measure), $L_1$ (the $L_1$ distance measure), $L_2$ (the $L_2$ distance measure), and cos (the cosine similarity measure).

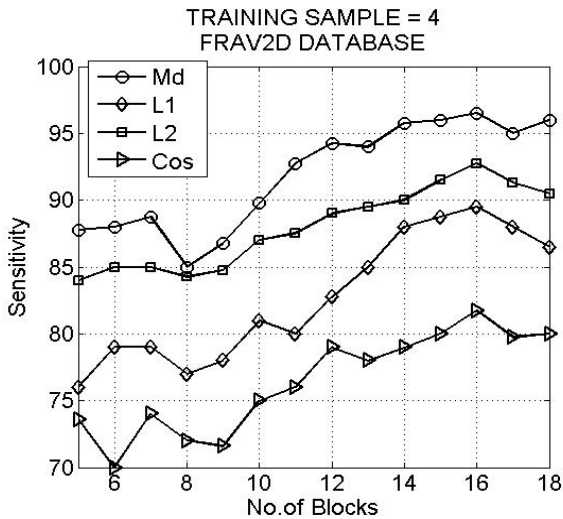

Fig. 13. Sensitivity measure of the Gabor based PHMM method on the FRAV2D database, using the four different similarity measures.

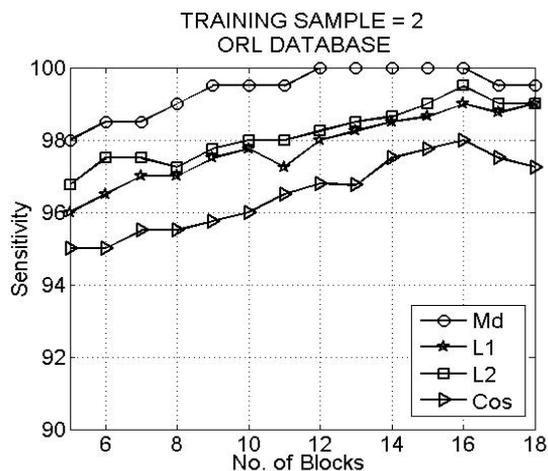

Fig. 14. Sensitivity measure of the Gabor based PHMM method on the ORL database, using the four different similarity measures.

Experimental results show that best results are achieved at block size of 16×16 pixels with an overlap of 75% in each direction. The Mahalanobis distance measure performs the best, followed in the order by the $L_1$ distance measure, the $L_2$ distance measure, and the cosine similarity measure. The reason for such an ordering is that the Mahalanobis distance measure counteracts the fact that $L_1$ and $L_2$ distance measures in the space weight preferentially for low frequencies. As the $L_2$ distance measure weights more the low frequencies than $L_1$ does, the $L_1$ distance measure should perform better than the $L_2$ distance measure, a conjecture validated by our experiments. The cosine similarity measure does not compensate the low frequency preference, and it performs the worst among all the measures. The experimental results show that one should use the Mahalanobis distance measure for the following comparative assessment of different face recognition methods.

As the Mahalanobis distance measure performs the best among all the four similarity measure, with. So the negative face recognition performance i.e. the specificity of the Gabor based PHMM method; with the same dataset, considering block size as 16×16 pixels and an overlap of 75% in each direction using the Mahalanobis distance similarity measures on the FERET, FRAV2D and ORL database respectively. The negative face recognition performance i.e. the specificity of the Gabor based PHMM method using the Mahalanobis distance similarity measure is shown in Fig. 15.

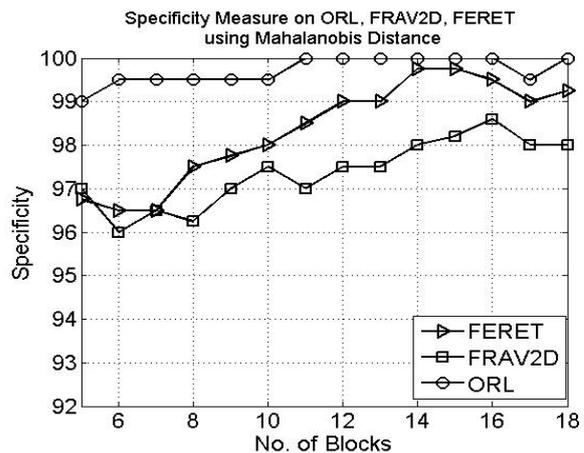

**Fig. 15**. Specificity measure of the Gabor based PHMM method on the FERET, FRAV2D, and ORL database, using the $M_d$ measure.

Table 7. Comparison of face recognition accuracy of the proposed method with wavelet/SHMM on the FERET Database.

| Method | Percentage(%) |
| --- | --- |
| DWT/HMM [27] | 42.9 |
| GWT/HMM [27] | 65.2 |
| **Proposed Method** | **91.5** |

*VIII. CONCLUSION*

In this paper, a hybrid approach is developed for face identification using PHMM together with GWT. Analysis has been carried out of the benefits of using GWT along with PHMM for face recognition. Also the effect of block size has been studied on the PHMM and therefore on the recognition accuracy. A new scanning technique for better feature selection has been applied to the PHMM. Here the use of most informative pixels of the GWT image instead of the whole image makes the HMM training and testing faster than the previous methods. The experimental result shows that use of Mahalanobis distance ($M_d$) measure classifier further enhances recognition accuracy and reduces the computational cost. As the $M_d$ classifier measures the distance between the optimal sequence path of the testing images with the optimal sequence path of the training classes only, instead of optimal sequence path of each of the training images, which, reduces the computational cost and makes this method practical for real-time applications.

ACKNOWLEDGEMENT

Authors are thankful to a major project entitled "Design and Development of Facial Thermogram Technology for Biometric Security System," funded by University Grants Commission (UGC),India and "DST-PURSE Programme" and CMATER and SRUVM project at Department of Computer Science and Engineering, Jadavpur University, Kolkata -700 032 India for providing necessary infrastructure to conduct experiments relating to this work.